\documentclass{article}

\usepackage{iclr2027_conference,times}
\usepackage{hyperref}
\usepackage{url}
\usepackage{amsmath,amssymb}
\usepackage{booktabs}
\usepackage{graphicx}
\hypersetup{hidelinks}

\newcommand{\method}{\textsc{OnlineQAT}}
\newcommand{\student}{\pi_{\theta}^{q}}
\newcommand{\teacher}{\pi_{T}}
\title{\method: On-Policy Distillation for Ultra-Low-Bit Large Language Models}

\author{Wenjun Wang$^{1,*}$ \quad Heng Li$^{2,*}$ \quad Yanggan Gu$^{1,*}$ \quad Hongxia Yang$^{1,3}$ \\
{\small $^{1}$The Hong Kong Polytechnic University \quad $^{2}$Sun Yat-sen University} \\
{\small $^{3}$PolyU-Daya Bay Technology and Innovation Research Institute \quad $^{*}$Equal contribution.}}

\iclrfinalcopy

\begin{document}

\maketitle
\lhead{Preprint. Work in progress.}

\begin{abstract}
Quantization-aware training (QAT) can recover much of the accuracy lost when large language models are compressed below four bits. Existing recovery stages, however, are commonly optimized on fixed completions or teacher-generated answers, whereas the deployed quantized model conditions on prefixes generated by itself. Quantization errors can therefore move the model into states that are absent from offline recovery data. We introduce \method{}, a two-stage framework that first obtains a usable low-bit initialization through block-wise QAT and then performs on-policy distillation (OPD) on student-generated responses. At each visited prefix, a frozen full-precision teacher provides a sampled reverse-KL training signal. On Qwen3-1.7B, \method{} obtains the best average among the compared quantized methods: 57.28 at W3A16 and 32.52 at W2A16, improving over ReasoningQAT by 2.90 and 0.44 points, respectively. The results suggest that student-visited states provide a useful recovery signal beyond fixed-completion training, particularly at three bits.
\end{abstract}

\section{Introduction}

Weight-only quantization reduces the memory footprint and bandwidth cost of large language models (LLMs), but two- and three-bit weights still incur a substantial accuracy loss. For reasoning models, a small perturbation to an early token can also change the states encountered throughout the remaining generation. Recovering useful behavior at such low precision therefore remains an open problem.

BitDistiller, EfficientQAT, and ReasoningQAT recover low-bit models through QAT and distillation~\citep{du2024bitdistiller,chen2024efficientqat,okoshi2026towards}. We use ReasoningQAT as our strongest offline baseline. Its recovery stage evaluates the student and teacher on gold completions and therefore visits reference prefixes. At inference time, however, the quantized model conditions on its own previous tokens. Once quantization changes an early decision, later prefixes can leave the distribution covered by the fixed-completion objective. Figure~\ref{fig:prefix} illustrates this distinction: offline recovery and deployment use different prefix distributions, whereas OPD trains on the prefixes visited by the deployed student.

\begin{center}
\begin{minipage}{\linewidth}
  \includegraphics[width=\linewidth]{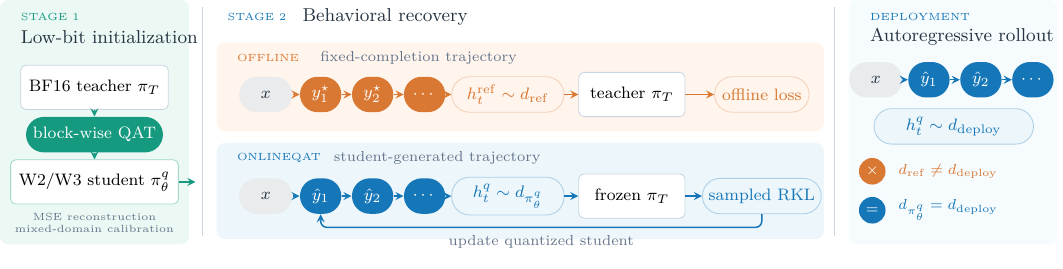}
  \refstepcounter{figure}\label{fig:prefix}
  \small Figure~\thefigure: Stage~1 produces a W2/W3 student. In Stage~2, offline recovery queries the teacher on fixed-completion prefixes, while deployment follows student-generated prefixes. OnlineQAT queries the frozen teacher on the student trajectory and applies sampled reverse KL, aligning the training and deployment states.
\end{minipage}
\end{center}

To address this mismatch, we propose \method{}, a two-stage recovery framework. Stage~1 performs block-wise QAT on mixed-domain calibration data, producing a student that can generate meaningful trajectories. Stage~2 samples responses from this student and uses a frozen full-precision teacher to provide a sampled reverse-KL signal at the visited prefixes. On Qwen3-1.7B, \method{} improves the six-task average over ReasoningQAT by 2.90 points at W3A16 and 0.44 points at W2A16, giving the best average among the compared methods in both settings. The W3A16 gain is broad, whereas the smaller W2A16 gain remains task-dependent.

\section{Related Work}

\paragraph{Ultra-low-bit QAT.}
QAT adapts model or quantization parameters while simulating low-precision weights. BitDistiller combines asymmetric quantization with confidence-aware self-distillation for sub-4-bit LLMs~\citep{du2024bitdistiller}. EfficientQAT introduces block-wise all-parameter training and a lightweight end-to-end quantization-parameter stage~\citep{chen2024efficientqat}. ReasoningQAT targets reasoning models with mixed-domain block-wise calibration followed by teacher-guided reward rectification and forward-KL distillation~\citep{okoshi2026towards}. These methods establish effective low-bit initialization and recovery recipes. When recovery is evaluated on fixed calibration sequences or reference completions, however, the objective does not cover prefixes induced by the deployed quantized policy. \method{} retains block-wise QAT but moves end-to-end recovery to those student-visited prefixes.

\paragraph{On-policy distillation.}
On-policy distillation trains a student on its own sampled sequences while querying a teacher at the visited states. GKD formalizes this idea for language-model distillation and shows that student-generated outputs can reduce the mismatch between training and generation~\citep{agarwal2023gkd}. Recent analysis studies when OPD succeeds and attributes its progress to alignment on high-probability tokens at student-visited states~\citep{li2026rethinkingopd}; large-scale systems such as Nemotron 3 Ultra further use multi-teacher OPD during post-training~\citep{nvidia2026nemotron}. Our setting is different: teacher and student share the same architecture, and their gap is created by ultra-low-bit weight quantization. We study whether OPD can recover this precision-induced gap.

\section{Method}

\method{} separates low-bit recovery into local initialization and end-to-end behavioral recovery. Let $\teacher$ be the frozen full-precision model and $\student$ its quantized counterpart. Stage~1 reconstructs one transformer block at a time. Stage~2 samples $\hat y$ from $\student$ and evaluates both models at student-generated prefixes $h_t=(x,\hat y_{<t})$, rather than the fixed prefixes $(x,y_{<t})$ used by offline distillation.

\subsection{Stage 1: block-wise low-bit initialization}

Following EfficientQAT and ReasoningQAT~\citep{chen2024efficientqat,okoshi2026towards}, Stage~1 uses sequential block-wise QAT on calibration data drawn from pre-training and reasoning corpora. Let $B_l$ and $B_l^q$ denote the full-precision and fake-quantized versions of block $l$, and let $H_l$ and $H_l^q$ be their respective input streams after the preceding blocks. For a sequence of $T$ tokens and hidden width $C$, we optimize
\begin{equation}
\mathcal L_{\mathrm{block}}^{(l)}=
\frac{1}{TC}\left\|B_l^q(H_l^q)-B_l(H_l)\right\|_F^2.
\label{eq:block}
\end{equation}
The fake quantizer uses asymmetric affine weight quantization with group size 128. We update the block's master weights, per-group scales and zero-points, and RMSNorm weights through the straight-through estimator, then propagate both streams to the next block. All end-to-end methods start from the resulting checkpoint.

\subsection{Stage 2: on-policy reverse-KL recovery}

For each prompt $x\sim\mathcal X$, the quantized student samples a response $\hat y\sim\pi_{\theta,\tau}^{q}(\cdot\mid x)$ at temperature $\tau$. At prefix $h_t=(x,\hat y_{<t})$, let $p_{\theta,t}^{\tau}$ and $p_{T,t}^{\tau}$ denote the temperature-scaled student and teacher distributions. The target objective is the reverse KL averaged over response tokens:
\begin{equation}
\mathcal L_{\mathrm{online}}(\theta)=
\mathbb E_{x\sim\mathcal X,\,\hat y\sim\pi_{\theta,\tau}^{q}}
\left[\frac{1}{|\hat y|}\sum_{t=1}^{|\hat y|}
D_{\mathrm{KL}}\!\left(
p_{\theta,t}^{\tau}\,\|\,p_{T,t}^{\tau}
\right)\right].
\label{eq:online}
\end{equation}
For the sampled token $a_t=\hat y_t$, a one-sample policy-gradient estimator uses the stop-gradient log-ratio
\begin{equation}
A_t^{\mathrm{RKL}}=
\operatorname{sg}\!\left[
\log p_{\theta,t}^{\tau}(a_t)-
\log p_{T,t}^{\tau}(a_t)
\right],
\label{eq:advantage}
\end{equation}
where $\operatorname{sg}$ stops gradients through the log-ratio. The training loss is
\begin{equation}
\mathcal L_{\mathrm{OPD}}(\theta)=
\frac{1}{N}\sum_{t\in\mathrm{response}}
A_t^{\mathrm{RKL}}\log p_{\theta,t}^{\tau}(a_t),
\label{eq:opd}
\end{equation}
with $N$ response tokens in the batch. The constant $+1$ term in the reverse-KL policy gradient is omitted because its expectation is zero. Minimizing Eq.~\ref{eq:opd} raises the probability of a sampled token when the teacher assigns it higher probability than the student, and lowers it in the opposite case. We use $\tau=0.6$ and no cross-entropy term.

\section{Experiments}

\subsection{Setup}

\paragraph{Benchmarks.}
We report GSM8K, MATH-500, MMLU-Redux (MMLU-R), IFEval, LiveCodeBench (LCB), and GPQA-Diamond (GPQA-D), covering mathematical reasoning, knowledge, instruction following, and code generation. AVG is their unweighted mean.

\paragraph{Model and quantization.}
We evaluate Qwen3-1.7B in thinking mode under W3A16 and W2A16. Transformer linear weights use asymmetric affine quantization along the input dimension with group size 128 and integer ranges $[0,7]$ and $[0,3]$. Activations, embeddings, and the LM head remain floating point; the teacher and zero-points are frozen during Stage~2.

\paragraph{Stage~1 calibration.}
We use 4,096 sequences of 2,048 tokens from 80\% OpenThoughts and 20\% FineWeb. Each block is trained for two epochs with batch size 2 and AdamW. Quantization parameters use learning rate $10^{-4}$; W3A16 and W2A16 master weights use $10^{-5}$ and $2\!\times\!10^{-5}$. All recovery methods start from the same Stage~1 checkpoint within each bit width.

\paragraph{Baselines and Stage~2.}
\textbf{Stg1} has no end-to-end recovery. \textbf{KD} is a fixed-prefix distillation baseline with $0.2\mathcal L_{\mathrm{CE}}+\mathcal L_{\mathrm{JSD}}$. \textbf{ReasoningQAT} follows Okoshi et al.~\citep{okoshi2026towards}, using $0.2\mathcal L_{\mathrm{RSFT}}+\mathcal L_{\mathrm{FKL}}$ on gold completions, where the forward KL retains the teacher's top-20 logits. It trains for 512 W3A16 or 1,536 W2A16 steps. \textbf{OnlineQAT} samples from 32,757 OpenThoughts prompts at temperature 0.6 and uses no supervised term. Sequences contain at most 8,192 total tokens and stop at EOS; only response tokens incur loss. All Stage~2 runs use AdamW on eight GPUs with effective batch 64. OnlineQAT trains W3A16 for 40 steps at $5\times10^{-6}$. W2A16 warms up for 30 steps from $2\times10^{-7}$ to $2\times10^{-6}$, followed by 120 steps at $2\times10^{-6}$.

\begin{table}[t]
\caption{Qwen3-1.7B results. ReasoningQAT denotes the method of \citet{okoshi2026towards}. All entries are percentages; AVG is the unweighted mean. Bold marks the best quantized result within each bit width.}
\label{tab:main}
\centering
\scriptsize
\setlength{\tabcolsep}{3.1pt}
\begin{tabular}{lrrrrrrr}
\toprule
Method & GSM8K & MMLU-R & MATH & IFEval & LCB & GPQA-D & AVG \\
\midrule
BF16 (Think) & 89.9 & 73.9 & 93.4 & 72.5 & 33.2 & 40.1 & 67.17 \\
\midrule
W3-Stg1 & 84.2 & 65.3 & 64.8 & 57.3 & 22.3 & 18.2 & 52.02 \\
W3-KD & 85.4 & 66.0 & 67.4 & 58.3 & 26.7 & 18.2 & 53.67 \\
W3-ReasoningQAT & 86.0 & 65.8 & 71.6 & \textbf{60.3} & 24.4 & 18.2 & 54.38 \\
\textbf{W3-OnlineQAT} & \textbf{86.3} & \textbf{68.0} & \textbf{73.2} & 59.5 & \textbf{27.4} & \textbf{29.3} & \textbf{57.28} \\
\midrule
W2-Stg1 & 24.40 & 17.20 & 21.80 & 33.20 & 0.50 & 11.60 & 18.12 \\
W2-KD & 55.00 & 38.40 & 31.40 & \textbf{41.30} & 4.10 & 14.10 & 30.72 \\
W2-ReasoningQAT & 56.60 & 43.80 & \textbf{37.80} & 33.80 & \textbf{5.30} & 15.20 & 32.08 \\
\textbf{W2-OnlineQAT} & \textbf{59.82} & \textbf{46.33} & 36.00 & 32.49 & 4.83 & \textbf{15.66} & \textbf{32.52} \\
\bottomrule
\end{tabular}
\end{table}

\subsection{Main results}

\paragraph{W3A16 recovery.}
\method{} averages 57.28, exceeding ReasoningQAT by 2.90 points and KD by 3.62 points. Relative to ReasoningQAT, it improves GSM8K, MMLU-R, MATH-500, LiveCodeBench, and GPQA-Diamond; the largest changes are $+3.0$ on LiveCodeBench and $+11.1$ on GPQA-Diamond. IFEval decreases by 0.8 points.

\paragraph{W2A16 recovery.}
At W2A16, \method{} improves the ReasoningQAT average from 32.08 to 32.52. It gains on GSM8K, MMLU-R, and GPQA-Diamond but trails on MATH-500, IFEval, and LiveCodeBench; the improvement is therefore modest and task-dependent.

\paragraph{Scope.}
Relative to ReasoningQAT, OnlineQAT changes both the prefix distribution and KL direction, so the comparison does not isolate either factor. The study covers one model without multiple training seeds or an offline reverse-KL control, and OPD incurs rollout generation. Table~\ref{tab:main} therefore supports an empirical comparison, not a causal or compute-efficiency claim.

\section{Conclusion}

\method{} retains the mixed-domain block-wise initialization of reasoning-oriented QAT and replaces fixed-sequence recovery with sampled-token OPD. On Qwen3-1.7B, this second stage improves the six-task average over ReasoningQAT by 2.90 points at W3A16 and 0.44 points at W2A16. These initial results motivate a broader evaluation of on-policy recovery for ultra-low-bit language models.

\begingroup
\scriptsize
\setlength{\bibsep}{0pt}
\bibliography{references}
\bibliographystyle{iclr2027_conference}
\endgroup

\end{document}